\documentclass{article}

\usepackage{microtype}
\usepackage{graphicx}
\usepackage{booktabs} % for professional tables

\usepackage{hyperref}

\usepackage[accepted]{icml2021}

\icmltitlerunning{A Policy Efficient Reduction Approach to Convex Constrained Deep Reinforcement Learning}

\usepackage{amsmath,amsfonts,bm}

\def\eqref#1{equation~\ref{#1}}
\def\1{\bm{1}}

\def\vzero{{\bm{0}}}
\def\vone{{\bm{1}}}
\def\vmu{{\bm{\mu}}}
\def\vtheta{{\bm{\theta}}}

\def\vc{{\bm{c}}}

\def\ve{{\bm{e}}}

\def\vq{{\bm{q}}}

\def\vs{{\bm{s}}}

\def\vu{{\bm{u}}}
\def\vv{{\bm{v}}}

\def\vx{{\bm{x}}}
\def\vy{{\bm{y}}}
\def\vz{{\bm{z}}}

\DeclareMathAlphabet{\mathsfit}{\encodingdefault}{\sfdefault}{m}{sl}
\SetMathAlphabet{\mathsfit}{bold}{\encodingdefault}{\sfdefault}{bx}{n}

\def\sC{{\mathbb{C}}}

\def\sP{{\mathbb{P}}}

\def\sS{{\mathbb{S}}}

\def\sV{{\mathbb{V}}}

\newcommand{\E}{\mathbb{E}}

\newcommand{\R}{\mathbb{R}}

\DeclareMathOperator*{\argmax}{arg\,max}
\DeclareMathOperator*{\argmin}{arg\,min}

\usepackage{hyperref}
\usepackage{url}

\usepackage{amsmath}  % for "align" environment
\usepackage{amsthm}   % for "proof" environment
\usepackage{amsfonts} % for \mathbb{}
\usepackage{dsfont}   % for indicator function font
\newtheorem{example}{Example}
\newtheorem{theorem}{Theorem}[section]

\newtheorem{corollary}{Corollary}[theorem] % Counter reset every time a new theorem environment is used.
\newtheorem{lemma}[theorem]{Lemma} % Re-use the counter as the theorem environment.
\newtheorem{proposition}[theorem]{Proposition} % 
\usepackage{algorithm} % for algorithm env
\usepackage{algorithmic} % for algorithm env
\renewcommand{\algorithmiccomment}[1]{\bgroup\hfill//~#1\egroup} % for comment in algorithm env

\usepackage{amssymb} % cross and tick % https://tex.stackexchange.com/questions/42619/x-mark-to-match-checkmark
\usepackage{pifont} % cross and tick
\newcommand{\cmark}{\ding{51}} % cross and tick
\newcommand{\xmark}{\ding{55}} % cross and tick
\usepackage{array} % using m{5em} in table
\newcolumntype{L}[1]{>{\raggedright\arraybackslash}m{#1}}  %% left aligned
\newcolumntype{C}[1]{>{\centering\arraybackslash}m{#1}} % for center and middle align in tabular
\newcolumntype{R}[1]{>{\raggedleft\arraybackslash}m{#1}}  %% right aligned
\usepackage{tablefootnote} % tablefootnote

\usepackage{graphicx}

\usepackage{subcaption} % subfigure environment

\usepackage{printlen} % to print line width: \printlength\textwidth

\def\sC{{\mathcal{C}}}

\def\sP{{\mathcal{P}}}

\def\sS{{\mathcal{S}}}

\def\sV{{\mathcal{V}}}

\def\proj{{\mathtt{Proj}}} % projection operator
\def\conv{{\mathtt{conv}}} % convex hull
\def\dist{{\mathtt{dist}}} % convex hull
\def\aff{{\mathtt{aff}}} % affine hull
\def\err{{\mathtt{err}}} % error

\def\valpha{{\boldsymbol{\alpha}}}

\def\vtheta{{\boldsymbol{\theta}}}

\def\vlambda{{\boldsymbol{\lambda}}}
\def\vmu{{\boldsymbol{\mu}}}

\def\vomega{{\boldsymbol{\omega}}}

\begin{document}

\twocolumn[
\icmltitle{A Policy Efficient Reduction Approach to Convex \\ Constrained Deep Reinforcement Learning}

% Submission and Formatting Instructions for \\
%   International Conference on Machine Learning (ICML 2021)}




\begin{icmlauthorlist}
\icmlauthor{Tianchi Cai}{ant}
\icmlauthor{Wenpeng Zhang}{ant_bj}
\icmlauthor{Lihong Gu}{ant}
\icmlauthor{Xiaodong Zeng}{ant}
\icmlauthor{Jinjie Gu}{ant}
\end{icmlauthorlist}
% Ant Group

\icmlaffiliation{ant}{Ant Group, Hangzhou, China.}
\icmlaffiliation{ant_bj}{Ant Group, Beijing, China}

% \icmlcorrespondingauthor{Tianchi Cai}{tianchi.ctc@antgroup.com}
% \icmlcorrespondingauthor{}{}

\icmlkeywords{Machine Learning, ICML}

\vskip 0.3in
]



\printAffiliationsAndNotice{}  % leave blank if no need to mention equal contribution

\begin{abstract}
Although well-established in general reinforcement learning (RL), value-based methods are rarely explored in constrained RL (CRL) for their incapability of finding policies that can randomize among multiple actions. To apply value-based methods to CRL, a recent groundbreaking line of game-theoretic approaches uses the mixed policy that randomizes among a set of carefully generated policies to converge to the desired constraint-satisfying policy. However, these approaches require storing a large set of policies, which is not policy efficient, and may incur prohibitive memory costs in constrained deep RL. To address this problem, we propose an alternative approach. Our approach first reformulates the CRL to an equivalent distance optimization problem. With a specially designed linear optimization oracle, we derive a meta-algorithm that solves it using any off-the-shelf RL algorithm and any conditional gradient (CG) type algorithm as subroutines. We then propose a new variant of the CG-type algorithm, which generalizes the minimum norm point (MNP) method. The proposed method matches the convergence rate of the existing game-theoretic approaches and achieves the worst-case optimal policy efficiency.  The experiments on a navigation task show that our method reduces the memory costs by an order of magnitude, and meanwhile achieves better performance, demonstrating both its effectiveness and efficiency.
\end{abstract}

\section{Introduction}
% RL is commonly use to learn sequential decision making policies, optimize the behavior of an agent in an unknown environment. 
% to not only maximize a long-term reward function, but also
% optimize the behavior of an agent in 

\begin{table*}[!tbp] 
\caption{Comparison of different works. Time complexity (number of RL tasks solved) and policy efficiency (number of neural networks stored) are compared, when using any deep RL method to find an $\epsilon$-approximate policy to a convex constrained RL problem with $m$-dimensional measurement function.} \label{table_methods_comparison}
\centering
\begin{tabular}{lccc} \toprule
\multicolumn{1}{C{3.2cm}}{Work} &
\multicolumn{1}{C{2.8cm}}{Time complexity}  &  % 2.4cm
\multicolumn{1}{C{2.8cm}}{Policy efficiency} & % 2.6cm
\multicolumn{1}{C{3.6cm}}{No extra hyperparameters} \\
 \midrule
 \citet{le2019batch} & \xmark  & $O(1/\epsilon)$ & \xmark\\ 
 \citet{miryoosefi2019reinforcement} & $O(1/\epsilon)$  & $O(1/\epsilon)$ & \xmark\\ 
Ours (Vanilla CG) & $O(1/\epsilon)$  & $O(1/\epsilon)$ & \xmark \\ 
Ours (Modified MNP) & $O(1/\epsilon)$  & $\le m+1 $ & \cmark \\ 
 \bottomrule
\end{tabular}
\end{table*}

% the desired behavior of the agent is naturally described using constraints
When applying reinforcement learning (RL) to many real-world tasks, it is inevitable to impose constraints to regulate the behavior of the resulting policy. Examples include adding risk constraints to avoid damaging expensive robotics \cite{blackmore2011chance,ono2015chance}, placing safety and comfort constraints on autonomous driving \cite{lefevre2015learning,shalev2016safe,isele2018safe,chen2019autonomous}, and introducing diversity constraints to encourage explorations \cite{hong2018diversity,miryoosefi2019reinforcement}. In general, such problems of learning desired policies under constraints can be cast into the  constrained reinforcement learning (CRL) formalism.

% the problem of learning to optimize the behavior of an agent under constraint is well known as  constrained reinforcement learning (CRL) problem.

% TODO @tianchi model-free or model-based.
As is well acknowledged, model-free RL methods can be classified into two major categories, i.e., value-based and policy-based \cite{sutton2018reinforcement}. However, compared with the large volume of literature studying value-based methods in the general RL setting, they are rarely investigated in the CRL setting. This somehow surprising phenomenon has its root cause that in CRL, a constraint-satisfying policy may require delicate randomization between different behaviors, and hence selecting multiple actions with specific probabilities is necessary (cf. Example \ref{why-value-based-fails}). Most value-based algorithms such as Q-learning \cite{sutton2018reinforcement}, DQN \cite{mnih2013playing}, and their variants \cite{van2015deep,wang2016dueling,lillicrap2015continuous,fujimoto2018addressing,barth2018distributed} may fail to find any constraint-satisfying policy in CRL. Therefore, the CRL literature traditionally merely focuses on policy-based methods  \cite{paternain2019constrained,tessler2018reward,achiam2017constrained,chow2017risk,chow2014algorithms}. Recently, value-based algorithms have achieved state-of-the-art performance in various RL tasks \cite{van2015deep,wang2016dueling,fujimoto2018addressing,barth2018distributed}. It is thus tempting to consider whether it is possible to solve CRL problems with value-based algorithms. 

A new line of research derived from the game-theoretic perspective has made a breakthrough in this direction \cite{le2019batch,miryoosefi2019reinforcement}. This line of game-theoretic approaches reformulates the CRL problem as a two-player zero-sum repeated game and solves it with no-regret online learning. In each round, one player who uses an online learning algorithm plays against the other player who uses an RL algorithm that finds a policy maximizing the value of the current game. This policy found by the RL player is then stored. It can be shown that after certain rounds, the \textit{mixed policy} that uniformly randomly selects one of the found policies converges to the desired constraint-satisfying policy. However, storing all the policies found by the RL player is not \textit{policy efficient} and may incur very high memory costs. In particular, when deep RL methods are utilized, even on some simple tasks, these game-theoretic approaches need to store dozens to hundreds of neural networks to find a constraint-satisfying policy (cf. Section \ref{experiment_sec}). In theory, to obtain an $\epsilon$-approximate policy in CRL, these game-theoretic approaches require storing $O(1/\epsilon)$ many policies, which is a consequence of their reliance on no-regret online learning \cite{freund1999adaptive,abernethy2011blackwell,hazan201210}. Given the high memory costs, the policy inefficiency of these game-theoretic approaches makes them impractical to work with deep RL methods.

To improve policy efficiency, we propose a novel \textit{vector space reduction approach} to solve the CRL problems. Instead of the game-theoretic perspective, we reduce the CRL problem over a policy space as an equivalent distance minimization problem over a vector space. We then show that this distance minimization problem can be solved by a specially designed conditional gradient (CG) algorithm, whose linear optimization oracle is constructed using an RL algorithm. Consequently, this reduction yields a meta-algorithm, which can be instantiated by any variant of CG and any off-the-shelf RL method. Specifically, in each iteration, the RL algorithm finds a policy, and this policy is stored; the mixed policy that selects all found policies with appropriate weights (e.g., step sizes) converges to a desired constraint-satisfying policy. The main benefit of our reduction approach is that it substitutes the no-regret online learning techniques with the CG-type methods, and thus it is not necessary to store all found policies. However, since the step sizes of the vanilla CG are non-zero, directly applying it assigns non-zero weights to all found policies and does not improve policy efficiency.

To this end, we propose a new algorithm, which achieves optimal policy efficiency, based on a variant of CG called the minimum norm point (MNP) method \cite{wolfe1976finding}. We extend the vanilla MNP to solve a more general problem, where the distance function to a convex set is minimized over a compact convex set. Inspired by the minor cycle technique \cite{wolfe1976finding} in MNP, our modified MNP method reassigns the weights of all found policies and maintains an active set, which only contains policies with non-zero weights. After the weight adjustment, policies with weight zero are eliminated from the active set immediately to cut the memory costs. To solve CRL problems with $m$-dimensional \textit{measurement vectors}, our method stores no more than $m+1$ policies throughout the learning process. Notably, this constant is shown to be worst-case optimal. Moreover, with a carefully refined analysis, our method solves the general problem with a faster convergence guarantee than the MNP method. To achieve an $\epsilon$-approximate solution in an $m$-dimensional space, our method improves the convergence from $O(m/\epsilon)$ \cite{chakrabarty2014provable} to a tighter $O(1/\epsilon)$, with the same memory cost (details in Table \ref{table_methods_comparison}). We compared our method with the game-theoretic approach \cite{miryoosefi2019reinforcement} in a navigation task using different RL methods to construct the oracle. In cases of both tabular RL and deep RL, our method demonstrates superior performance and policy efficiency. In particular, in deep RL cases, our method even reduces the memory costs by an order of magnitude. In summary, our approach enables efficiently utilizing value-based RL methods to solve CRL, and the improved policy efficiency (worst-case optimal) makes it especially appealing to applications using deep RL methods.

\section{Background} \label{sec_c2rl}
%
% We first introduce notation. A \textit{vector-valued Markov decision process} is a tuple $\{S, A, \beta, P, \vc\}$, where $S$ is the set of states, $A$ is the set of actions, $\beta\colon S \to [0,1]$ is the initial state distribution, $P\colon S\ \times A \times S \to [0,1]$ is the transition probability function (where $P(s'| s, a)$ is the probability of transitioning to state $s'$ after taking action $a$ in state $s$) and $\vc\colon S \times A \times S \to \R^m$ is an m-dimensional measurement vector function.

% . At the start of each episode, an initial state $s_0$ is drawn following the distribution $\beta$. Then, at each step $t=0,1,\dots$, the agent observes a state $s_t\in S$ and makes a decision to take an action $a_t$. After $a_t$ is chosen, at the next observation the state evolves to state $s_{t+1} \in S$ with probability $P(s_{t+1} | s_t, a_t)$. However, instead of a scalar reward, in our setting, the agent receives an $m$-dimensional vector $\vc_t \in \R^m$ that may implicitly contain measurements of reward, risk or violation of other constraints. The episode ends after a certain number of steps, called the horizon, or when a terminate state is reached.

% We first introduce notation. 
A \textit{vector-valued Markov decision process} is defined as a tuple $\{S, A, P, \beta, \vc, \gamma\}$, where $S$ is a set of states $s\in S$, $A$ is a set of actions $a\in A$, $P$ is a transition probability function of the form $P(s_{t+1}| s_t, a_t)$ that describes the dynamics of the system, $\beta$ defines the initial state distribution $\beta(s_0)$, $\vc\colon S \times A \to \R^m$ is an $m$-dimensional \textit{measurement} function $\vc(s_t, a_t)$ that may measure reward, risk or other constraints, and $\gamma\in[0,1)$ is a discount factor.

% Most methods in constrained RL focuses on policy-based methods and finds 
Actions are typically selected according to some (stationary) policies. A policy $\pi$ maps states to probability distributions over actions, and $\pi(a_t|s_t)$ denotes the probability of selecting action $a_t$ in $s_t$. We assume that policies under consideration are selected from some candidate policy set $\Pi$. For example, in policy-based methods, $\Pi$ is usually the set of all stationary policies, and in value-based methods, $\Pi$ is typically the set of all deterministic policies. For a policy $\pi\in\Pi$, we define the \textit{long-term measurement} $J(\pi)$ as the expectation of the discounted cumulative measurements
\begin{align}
  J(\pi) := \E\left[\sum_{t=0}^{\infty} \gamma^t \vc(s_t, a_t) \middle| a_t \sim \pi(\cdot | s_t)\right],
\end{align}
where the expectation is over the described random process.

To enable utilizing value-based methods to solve CRL problems, we also consider \textit{mixed policies}, which are distributions over the candidate policies space $\Pi$. We define $\Delta(\Pi)$ to be the set of all mixed policies generated by $\Pi$. To execute a mixed policy $\mu \in \Delta(\Pi)$, at the start of an episode, we select a policy $\pi\sim\mu(\pi)$, and then execute $\pi$ for the entire episode. The long-term measurement of a mixed policy is defined accordingly:
\begin{align}
  J(\mu) := \E_{\pi \sim \mu}[J(\pi)] = \sum_{\pi \in \Pi} \mu(\pi) J(\pi).
\end{align}
In the following, we focus on the convex constrained RL problem, also known as the \textit{feasibility problem}, which generalizes inequality constraints to convex constraints \cite{miryoosefi2019reinforcement}. A \textit{feasibility problem} is specified by a closed and convex set $\Omega \in\R^m$. The goal is to find a policy whose long-term measurement lies inside $\Omega$.
\begin{align}
 \text{\ Find a policy }\mu \in\Delta(\Pi)\text{ such that } J(\mu) \in \Omega. \label{constrained_problem}
\end{align}
A policy is \textit{feasible} if it satisfies the constraint, and the problem is \textit{feasible} if a feasible policy exists. This formulation can potentially handle tasks that maximize one measurement (e.g., reward) under convex constraints. Such problems can be solved by performing a binary search over the maximum achievable reward value and at each iteration augmenting an inequality reward constraint (reward no less than the current iterated value) to the constraints.

Though both policy-based methods and value-based methods are well established in general RL, in the feasibility problem, the feasible policies may require choosing among multiple actions with specific probabilities, which is not satisfied by many value-based methods. We illustrate this difficulty with the following example.

\begin{example}\label{why-value-based-fails}
We consider the task of playing the Rock, Paper, Scissors game. For simplicity, we assume the environment randomly selects one of the three actions, and the game terminates after a fixed number of rounds. Let the measurement vector be the basis vectors in $\R^3$, indicating whether the agent won with each of the three actions, and the zero vector if tie or loss. Consider the feasibility problem specified by $\Omega = \{(i, j, k)\ | i,j,k\ge 1/9\}$, which requires the agent to win with each action with at least $1/9$ probability on expectation. It is obvious that the only feasible policy for this task is to select three actions with the same probability. However, most value-based methods calculate a scalar value for each state-action pair and select any action achieving the maximum value at the current state. Since value-based methods cannot specify the probability for choosing each state-action pair, they may fail to solve CRL problems.
\end{example}

One workaround is to use mixed policies. However, the main difficulty to use mixed policies is that when each policy is found by a deep RL method, the memory costs can be huge. To store such a mixed policy $\mu$, the neural networks corresponding to all policies with non-zero probability have to be stored. Hence the memory cost of storing $\mu$ is proportional to the cardinality of the subset of policies with non-zero weights in the candidate policy space, i.e., $|\{\pi|\pi\in \Pi, \mu(\pi)>0\}|$. Since a neural network may have billions to trillions of parameters \cite{brown2020language,fedus2021switch}, storing a large number of neural networks is impractical in many deep RL tasks. Therefore, we are interested in mixed policies that are \textit{policy efficient} and have a small cardinality of policies with non-zero weights. 

\section{A Vector Space Reduction Approach}

Our vector space reduction approach reformulates the original CRL problem over a policy space to an equivalent distance minimization problem over a vector space. The key is to construct a specific linear optimization oracle using any RL algorithm, which enables solving this distance minimization problem with any variant of the CG method. This reduction yields a meta-algorithm for the CRL problems, which can be instantiated by any CG method and any RL algorithm. We illustrate this with the vanilla CG method.

% modular Frank-Wolfe (CG) approach to the constrained RL problem, and several resulting methods using variants of CG. We then propose a novel variant by modifying Wolfe's Minimum Norm Point (MNP) method. We show this modified MNP method matches the $O(1/t)$ convergence rate of previous methods, and achieves the optimal sparsity of storing no more than $m+1$ policies. \answerTODO{based on the observation that the measurement vector lies inside a convex polytope. We reduce the crl to a convex optimization problem in $R^m$}

\subsection{Equivalent Distance Minimization Problem}
We first reformulate the feasibility problem to an equivalent distance minimization problem over the policy space.  For a closed and convex set $\Omega\in\R^m$, considering the problem of finding a mixed policy $\mu\in\Delta(\Pi)$, whose long-term measurement is closest to the target convex set,
\begin{align}
    \argmin_{\mu\in\Delta(\Pi)}  \frac{1}{2}\dist^2(J(\mu), \Omega) \label{lmo_problem},
\end{align}
where $\dist(\vx, \Omega) := ||\vx - \proj_\Omega(\vx)||$ is the Euclidean distance of $\vx$ to the set $\Omega$, and $\proj_\Omega(\vx) \in \argmin_{\vy \in \Omega} ||\vx - \vy||$ is the Euclidean projection of $\vx$ onto the set $\Omega$.

For this minimization problem, a policy $\mu^*\in\Delta(\Pi)$ is defined to be \textit{optimal} if it minimizes (\ref{lmo_problem}). Otherwise, the \textit{approximation error} of $\mu\in\Delta(\Pi)$ is defined as 
\begin{align}
\mathtt{err}(\mu) := \frac{1}{2}\dist^2(J(\mu), \Omega) - \frac{1}{2}\dist^2(J(\mu^*), \Omega).  \label{approximation_error} % \quad\quad \text{(Approximation Error)}
\end{align}
A policy is defined to be an $\epsilon$-approximate policy if its approximation error is no larger than $\epsilon$. 

When the CRL problem (\ref{constrained_problem}) is feasible, the equivalence of being optimal to (\ref{lmo_problem}) and being feasible to the CRL problem can be easily established. Since a feasible policy of the CRL problem lies inside $\Omega$, it minimizes the non-negative $\dist(\cdot, \Omega)$ function, and hence is optimal to (\ref{lmo_problem}). Vice versa, any optimal policy to (\ref{lmo_problem}) lies inside $\Omega$ and is a feasible policy to the CRL problem.

From a geometric perspective, let $J(\Pi) := \{J(\pi) | \pi \in \Pi\}$ denote the set of all long-term measurements achievable by policies in the candidate policy space $\Pi$. It is clear that 
\begin{align*}
    J(\Delta(\Pi)) &= \{J(\mu) | \mu \in \Delta(\Pi)\}\\
    &= \left\{\sum_{\pi\in\Pi}\mu(\pi)J(\pi)\middle |\sum_{\pi\in\Pi}\mu(\pi) =1, \mu(\pi)\ge 0\right\} %\subset \R^m
\end{align*}
is the convex hull of $J(\Pi)$, and hence is closed and compact. Therefore the distance minimization problem (\ref{lmo_problem}) over the policy space $\Delta(\Pi)$ is equivalent to the following distance minimization problem over a closed and convex set $J(\Delta(\Pi))\in\R^m$:
\begin{align}
    \argmin_{J(\mu)\in J(\Delta(\Pi))} \frac{1}{2} \dist^2(J(\mu), \Omega). \label{dmp}
\end{align}
If the CRL problem is feasible, then any $J(\mu^*)$ that minimizes this distance function over the convex set $J(\Delta(\Pi))$ finds a feasible policy $\mu^*$ to the original problem. Hence we have reduced the original CRL problem over a policy space $\Delta(\Pi)$ to an equivalent distance minimization problem (\ref{dmp}) over the closed and convex set $J(\Delta(\Pi))$ in a vector space.

\subsection{A Solution with Vanilla Conditional Gradient}

Since it is unclear how to project a policy to the implicitly defined set $J(\Delta(\Pi))$, this distance minimization problem (\ref{dmp}) is non-trivial. We overcome this difficulty by proposing a specially designed conditional gradient (CG) algorithm, where the linear optimization oracle used by the CG method is constructed using any off-the-shelf RL algorithm.

We briefly review the CG method. CG is a first-order method to minimize a convex function $f: \sC \mapsto \R$  over a compact and convex set $\sC$, using a linear optimization oracle \cite{frank1956algorithm}
\begin{align}
    \min_{\vx\in\sC} f(\vx)\  \text{ using }\  \mathtt{Oracle}(\vv) := \argmin_{\vs\in\sC} \vs^T\vv. \label{fw_statement}
\end{align}
In each iteration step $t$, the CG (Algorithm 3 in Appendix A.1) calculates the gradient at the current point $\nabla f(\vx)$, and invokes the linear optimization oracle to find an improving point $\vs \gets \mathtt{Oracle}(\nabla f(\vx))$. Then it updates the iterated point by taking a convex average of the current point $\vx$ and the improving point $\vx \gets (1-\eta_t)\vx + \eta_t\vs$ , where at step $t$, the step size is typically set to $\eta_t \gets \frac{2}{t+1}$ \cite{jaggi2013revisiting}.
\begin{algorithm}[bt]
\caption{Solve a CRL Problem with Vanilla CG} \label{algo_fw}
\textbf{Input}: $\mathtt{RL\_oracle}(\cdot)$, learning rate $\{\eta_t\in[0,1]\}_{t=1,\dots,T}$\\
\textbf{Initialize}: $\vx_0\in\R^m$

\begin{algorithmic}[1] %[1] enables line numbers
\FOR{t = 1, \dots, T}
\STATE $\pi_t, J(\pi_t) \gets \mathtt{RL\_oracle}(\vx_{t-1} - \proj_\Omega(\vx_{t-1}))$
\STATE $\vx_t \gets (1-\eta_{t})\vx_{t-1} + \eta_tJ(\pi_t)$, \\ 
$\mu_t \gets (1-\eta_{t})\mu_{t-1} + \eta_t\pi_t$ \COMMENT{A new policy is stored} \label{fw_algo_line2}
\ENDFOR
\STATE \textbf{return} $\vx_T, \mu_T$ \COMMENT{$\vx_T: \vx_T = J(\mu_T)$}
\end{algorithmic}
\end{algorithm}

We first calculate the gradient of the target function with respect to $\vx$. (1.1) of \citet{holmes1973smoothness} shows that the gradient of the function $\dist(\vx, \Omega)$ with respect to $\vx$ is $\nabla \dist(\vx, \Omega) = \mathtt{sgn}(\vx-\proj_\Omega (\vx))$, where $\mathtt{sgn}(\vx)=\frac{\vx}{||\vx||}$ if $\vx \neq \vzero$ else $\vzero$. Hence applying the chain rule, it is straightforward that
\begin{align*}
\nabla \frac{1}{2}\dist^2(\vx, \Omega) = \vx-\proj_\Omega (\vx).
\end{align*}
We construct the desired linear optimization oracle, denoted by $\mathtt{RL\_oracle}(\vlambda)$, such that for any $\vlambda\in\R^m$, it outputs a policy, together with the corresponding measurement vector
\begin{align}
    \pi, J(\pi) \gets \mathtt{RL\_oracle}(\vlambda),
\end{align}
satisfying $\pi \in \argmin_{u\in\Pi}\vlambda^T J(u)$. To construct the linear optimization oracle, in fact the improving policy $\pi$ can be found by using any off-the-shelf RL algorithms to solve a specific RL task. In particular, for any $\vlambda \in \R^m$, a policy that minimizes
\begin{align}
    \argmin_{\pi\in\Pi} \vlambda^T J(\pi) 
        &= \argmin_{\pi\in\Pi} \vlambda^T\E(\sum_{t=0}^\infty \gamma^t \vc_t)\\
        &= -\argmax_{\pi\in\Pi} \E(\sum_{t=0}^\infty \gamma^t (-\vlambda^T\vc_t)) \label{construct_rl_oracle},
\end{align}
is a policy that maximizes the scalar reward $r_t := (-\vlambda^T\vc_t)$ at each step. Therefore any reinforcement learning algorithm that maximizes this scalar reward $r$ finds an improving policy, and the RL algorithm that best suits the underlying problem can be used to find an improving policy $\pi$. 

Evaluating the measurement vector $J(\pi)$ is handy in online settings, where Monte Carlo simulations estimate $J(\pi)$ directly. In batch or offline settings, various off-policy evaluation methods, such as importance sampling \cite{precup2000eligibility,precup2001off} or doubly robust \cite{jiang2016doubly,dudik2011doubly}, can be used to estimate $J(\pi)$. 

With the linear optimization oracle $\mathtt{RL\_oracle}(\cdot)$ constructed using any RL method, the distance minimizing problem (\ref{dmp}) can be solved by any variant of the CG-type algorithm. When the vanilla CG algorithm is used, the resulting method is illustrated in Algorithm \ref{algo_fw}. In each iteration, the $\mathtt{RL\_oracle}(\cdot)$ is invoked once to find an improving policy $\pi_t$, together with its long-term measurement $J(\pi_t)$. Then, the current mixed policy $\mu_t$ is updated by selecting $\pi_t$ with weight $\eta_t$, and selecting any previously found policy $\pi_i \in \{\pi_1, \dots, \pi_{t-1}\}$ with weight $(1-\eta_t)\mu_{t-1}(\pi_i)$. The iterated point $\vx_t$ is updated in the same way, ensuring the invariance that $\vx_t = J(\mu_t)$. The convergence of the vanilla CG is well-established \cite{jaggi2013revisiting,lan2020first}, which readily implies the Algorithm \ref{algo_fw} converges in a sublinear $O(1/t)$ convergence rate. However, since the learning rates of vanilla CG is always non-zero, after $T$ iterations, all policies $\{\pi_1, ...\pi_T\}$ have non-zero weights to be selected in $\mu_T$. When the policies are found by deep RL methods, this requires storing $T$ neural networks, and is not policy efficient. We conclude that Algorithm \ref{algo_fw} matches the $O(1/t)$ convergence rate and $O(1/t)$ policy efficiency of the existing game-theoretic approaches.

\section{A Policy Efficient CG Approach}

Comparing with the game-theoretic approaches, our vector space reduction approach does not require storing all found policies. However, directly applying the vanilla CG method assigns non-zero weights to all found policies and does not improve policy efficiency. To improve policy efficiency, we propose a new CG-type method. Our method is based on a variant of CG called the minimum norm point (MNP) method \cite{wolfe1976finding}. We extend the MNP to solve a more general problem. When applying to the CRL problem, we show that our proposed method matches the $O(1/t)$ convergence rate and achieves an optimal policy efficiency.

\subsection{Minimum Norm Point Method} 

% \answerTODO{discuss of CG algos, literature review, explains MNP as background. More explanations on minor cycle. Propose our method, give convergence proofs (add the oracle approximation and binary search part).}

To find \textit{policy efficient} mixed policies, we turn to variants of CG-type algorithms, especially those that maintains an active set, and assign zero-weights to certain iterated points. When the target convex set is a singleton, a policy efficient solution can be readily found using Wolfe's method for Minimum Norm Point (MNP) over a polytope \cite{wolfe1976finding,de2018minimum}. 

When the target set is a singleton containing one point $\Omega = \{\vomega\}$, the distance minimization problem (\ref{dmp}) is simplified to finding a point in the polytope that is closest to $\vomega$
\begin{align}
    \argmin_{J(\mu)\in J(\Delta(\Pi))} \frac{1}{2} ||J(\mu) - \vomega ||^2,
\end{align}
which can be readily solved by Wolfe's method for finding Minimum Norm Point (MNP) in a polytope. % (MNP minimizes the Euclidean norm instead of the squared Euclidean norm, which is equivalent since we are interested in the policy that achieves the minimum.)

In MNP (Algorithm 4 in Appendix A.2), the loop in CG is called a \textit{major cycle}, and the convex averaging step is replaced by weight reassignment processes, called \textit{minor cycles}. MNP maintains an active set $S$, and the current iterated point is represented as a convex combination of points in $S$. 

Recall that for a set of points $S := \{\vc_0, \dots, \vc_k\}$, the affine hull $ \aff(S)$ is defined as 
\begin{align}
% \conv(S) &:= \{\sum_{i=0}^k\vmu_i\vc_i\ | \sum_{i=0}^k \vmu_i=1, \vmu\ge 0\}, \\ 
\aff(S) &:= \{\sum_{i=0}^k\vmu_i\vc_i\ | \sum_{i=0}^k\vmu_i =1, \vmu \in \R^k\}.
\end{align}
The convex hull $\conv(S)$ is defined similarly with an additional requirement that $\vmu \ge \vzero$ elementwise. The \textit{affine minimizer} is defined as $\argmin_{\vc\in\aff(S)} ||\vc||_2$. When a point $\vomega\in\R^m$ is treated as the origin, the affine minimizer with respect to $\vomega$ is $\vy := \argmin_{\vc\in\aff(S)} ||\vc - \vomega||_2$ and the affine minimizer property gives
\begin{align}
\forall \vv \in \aff(S), (\vv-\vy)^T (\vy-\vomega) = 0  \label{affine_minimizer_property}
\end{align}
In a major cycle, when $\Omega = \{\vomega\}$, we have $\proj_\Omega(\vx_{t-1}) = \vomega$ for all $\vx_{t-1}$. Hence, the MNP uses the oracle the same way as the CG algorithm. To minimize the size of the active set, the MNP repeatedly eliminates points from the active set using minor cycles. The minor cycles are executed until $S$ becomes a \textit{corral}, that is, its affine minimizer lies inside its convex hull. To maintain the corral property of active set $S$, in a minor cycle, let $\vy$ be the point of smallest norm in of the affine hull $\aff(S)$. If $\vy$ is in the relative interior of the convex hull $\conv(S)$, then the minor cycle is terminated. Otherwise, $\vx_{t-1}$ is updated to the nearest point to $\vy$ on the line segment $\conv(S) \cap [\vx_{t-1}, \vy]$. Thus $\vx_{t-1}$ is updated to a boundary point of $\conv(S)$, and any point, not on the face of $\conv(S)$ in which $\vx_{t-1}$ lies, is deleted. Note that singletons are always corrals, and hence the minor cycles terminate after a finite number of runs. After which $\vx_t$ is updated to the affine minimizer $\vy$ of the corral $S$.

The process $\mathtt{AffineMinimizer}_{\vomega_t}(S) $ returns $(\vy,\alpha)$ the affine minimizer $\vy$ of $S$ and $\alpha : \vy = \sum_{\vx\in S} \alpha(\vx)\vx$ is the coefficient expressing $\vy$ as an affine combination of points in $S$, where $\alpha(\vx)$ is the weight associated with $\vx$. The process $\mathtt{AffineMinimizer}_{\vomega_t}(S) $ can be straightforwardly implemented using linear algebra. \citet{wolfe1976finding} also provides a more efficient implementation that uses a triangular array representation of the active set. 

In the singleton case, the MNP solves the distance minimization problem (\ref{dmp}), and hence the CRL problem (\ref{constrained_problem}). Since the active set is a corral and hence is affinely independent, the number of policies stored is at most $m+1$ at any time. After $t$ major cycle steps, the MNP method is shown to converge linearly with a rate $O(e^{-\rho t})$ where $\rho$ is an constant determined by the polytope as defined in \citet{lacoste2015global}.

\begin{algorithm}[tb]
\caption{Solve a CRL Problem with Modified MNP} \label{algo_modified_mnp}
\textbf{Input}: $\mathtt{RL\_oracle}(\cdot)$, target set $\Omega$.\\
\textbf{Initialize}: current point $\vx_0=\vzero\in\R^m$, active set $S = \{\}$, active policy set $A=\{\}$, weight $\mu(\cdot)$ for policies in $A$.

\begin{algorithmic}[1] %[1] enables line numbers

\FOR[Major cycle]{t = 1, \dots, T}
\STATE $\vomega_t \gets \proj_\Omega (\vx_{t-1})$
\STATE $\pi_t, J(\pi_t) \gets \mathtt{RL\_oracle}(\vx_{t-1} - \vomega_t)$ \label{algo_mnp_line2}
% \IF{$J(\pi_t)$ not in $\conv(S)$}
\STATE $A \gets A \cup \{\pi_t\}, S \gets S \cup \{J(\pi_t)\}$
% \ENDIF
\WHILE[Minor cycle]{True} \label{minor_cycle_start}
  \STATE $\vy, \alpha \gets \mathtt{AffineMinimizer}_{\vomega_t}(S)$ \label{algo_distance_minimizing_affine_minimizer}
  \IF[$S$ is a corral]{$\forall \pi \in A, \alpha(\pi) > 0$}
  \STATE \textbf{break}
  \ELSE
    \STATE $\theta \gets \min_{\pi: \alpha(\pi)\le 0} \frac{\mu(\pi)}{\mu(\pi) - \alpha(\pi)}$
    \STATE $\vx_{t-1} \gets \theta\vy + (1-\theta) \vx_{t-1}$, \\$ \mu(\pi) \gets \theta\alpha(\pi) + (1-\theta) \mu(\pi)$ for all $\pi\in A$
    \STATE // Remove policies with weight zero
    \STATE $A \gets \{\pi\in A\ |\ \mu(\pi)>0\}$, \\ $S \gets \{J(\pi)\in S\ |\ \mu(\pi)>0\}$ \COMMENT{Save memory}
  \ENDIF
\ENDWHILE \label{minor_cycle_end}
\STATE $\vx_t \gets \vy, \mu(\pi) \gets \alpha(\pi)$ for all $\pi\in A$
\ENDFOR
\STATE \textbf{return} $\vx_T, \mu_T$ \COMMENT{$\vx_T: \vx_T = J(\mu_T)$}
\end{algorithmic}
\end{algorithm}

\subsection{Modified MNP and Theoretical Analysis}

To solve the general case where $\Omega$ may not be a singleton, we propose a modified MNP method. In the general non-singleton case, our target function is in fact not strongly-convex (Proposition \ref{not_stronly_convex}). We analyze the complexity of our modified MNP method, and improve from the previous $O(m/t)$ \cite{chakrabarty2014provable} to a tighter $O(1/t)$ convergence rate (Theorem \ref{convergence}). Moreover, we show that maintaining an active policy set of size $m+1$ is worst-case optimal (Theorem \ref{optimal_constant_general}). Therefore we conclude that the proposed modified MNP method matches the $O(1/t)$ convergence of the existing game-theoretic methods, and achieves an optimal policy efficiency of storing no more than $m+1$ policies.

As illustrated in Algorithm \ref{algo_modified_mnp}, we modify the MNP by adding a projection step into the major cycle (line \ref{algo_mnp_line2}). In each major cycle, the modified MNP minimizes the distance to a projected point $\vomega_t := \proj_\Omega(\vx_{t-1})$. Hence the resulting algorithm is equivalent to Wolfe's MNP method when $\Omega$ is a singleton, and otherwise, the oracle step calculates the gradient the same as the CG method. Intuitively, at each major step, if we are making a significant progress toward the projected point, then the distance to the convex set is decreased by at least the same amount. 

For non-singleton $\Omega$, in fact we cannot achieve the linear convergence as the singleton case. This is because in a non-singleton case, the target squared distance function is not strongly convex, which is a common assumption required for linear convergence.

Recall that a function $f$ over $S$ is defined to be strongly convex  \cite{boyd2004convex}, if there exists $m>0$, such that for all $\vu,\vv \in S$, $f$ satisfies
\begin{align}
    f(\vu)\ge f(\vv) + \nabla f(\vv)^T(\vu-\vv) + \frac{m}{2}||\vu-\vv||^2.
\end{align}
\begin{proposition} \label{not_stronly_convex}
For any convex set $\Omega$, the function $f(\vx) := \frac{1}{2} \dist^2(\vx, \Omega)$ is strongly convex if and only if $\Omega$ is a singleton. 
\end{proposition}
A proof is given in Appendix B.1. This proposition shows that the singleton case solved by MNP is the only case where the target function is strongly convex and linear convergence can be achieved. For general non-singleton $\Omega$, the linear convergence does not hold. To analyze the convergence of our modified MNP method, we first show that the approximation error strictly decreases between any two steps.

\begin{theorem} [Approximation Error Strictly Decreases] \label{thm_strictly_decrease} 
For each $t>1$ step, the $\vx_t, \mu_t$ found by Algorithm \ref{algo_modified_mnp} satisfies $\err(\mu_{t}) < \err(\mu_{t-1})$. That is, the measurement vectors of $\mu_{t}$ get strictly closer to the convex set $\Omega$.
\end{theorem}
A proof is provided in Appendix B.2. Given the approximation error strictly decreases, MNP can be shown to terminate finitely \cite{wolfe1976finding}. However, this finitely terminating property does not hold for our algorithm. Since a changed projected point $\omega$ may yield a lower distance for the same active set $S_t$, the active set may stay unchanged across major cycles (cf. Section \ref{experiment_sec}). We establish the convergence of the modified MNP method by the following theorem.
\begin{theorem} [Convergence in Approximation Error] \label{convergence} 
For any $t\ge 1$, the mixed policy $\mu_t$ found by the modified MNP method (Algorithm \ref{algo_modified_mnp}) satisfies
\begin{align}
    \err(\mu_t) \le 16Q^2/(t+2),
\end{align} 
where $Q := \max_{\pi\in\Pi} ||J(\pi)||$ is the maximum norm of a measurement vector. 
\end{theorem}

The proof is provided in Appendix B.3. In short, we define major cycle steps with at most one minor cycle as \textit{non-drop step}, which are "good" steps, and major cycle steps with more than one minor cycles as \text{drop steps}, which are "bad" steps. We show that in good steps, Algorithm \ref{algo_modified_mnp} is guaranteed to make enough progress. Though this does not hold for bad steps, we can bound the frequency of bad steps, and by Theorem \ref{thm_strictly_decrease}, bad steps still make progresses. Hence the convergence follows. The main techniques are based on \citet{chakrabarty2014provable}. However, since we give a tighter bound on the frequency of bad steps, we improves the convergence rate from their $O(m/t)$ to a tighter $O(1/t)$.

\begin{figure}[tbh] 
\begin{center}
\includegraphics[width=0.28\textwidth]{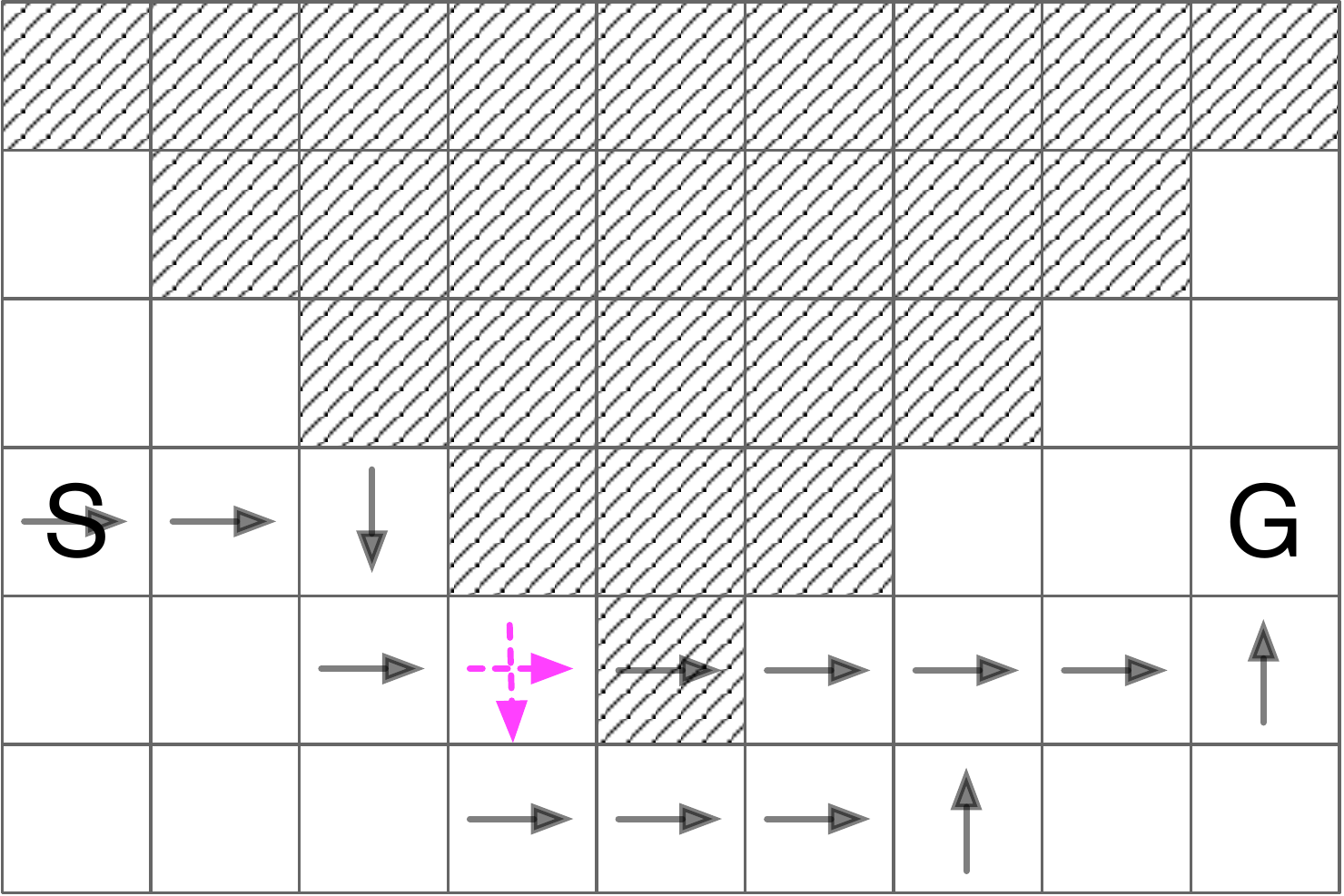}
\end{center}
\caption{The illustration of the navigation task. The agent needs to navigate from S to G, with no more than 11 steps, and no more than 0.5 steps in the grey region on expectation. This requires the agent to randomly choose between safer paths and shorter paths.} \label{figure_env}
% \captionsetup{belowskip=0pt}
% \vspace{-1.5em}
\end{figure}

% The proof is provided in Appendix \ref{convergence_proof}, which relies on the following two lemmas. 
% %
% \begin{lemma} \label{lemma_non_drop_steps}
% For a non-drop step in C2RL method, we have $\err(\mu^t) - \err(\mu^{t+1}) \ge \err^2(\mu^t)/8Q^2$. 
% \end{lemma}
% %
% Though this does not hold for drop steps, we can bound the frequency of drop steps by the following. 
% %
% \begin{lemma} \label{lemma_drop_steps}
% After $t$ major cycle steps of C2RL method, the number of drop steps is less than $t/2$.
% \end{lemma}
% %
% Since the approximation error strictly decreases (Thm. \ref{thm_strictly_decrease}), and in more than half of the major cycles steps, the C2RL method makes significantly progress. The Thm. (\ref{convergence}) can then be proved using an induction argument (Appendix \ref{convergence_proof}).

% \vspace{-0.5em}

We then discuss the policy efficiency of mixed policy for the CRL problem. We give a constructive proof in Appendix B.4 to show that to ensure convergence for RL algorithms whose candidate policy set are deterministic policies (e.g. DQN \cite{mnih2013playing}, DDPG \cite{lillicrap2015continuous} and variants \cite{van2015deep,wang2016dueling,fujimoto2018addressing,barth2018distributed}), storing $m+1$ policies is necessary in the worst case. 

\begin{figure*}[tb] 
\begin{center}
\includegraphics[width=\textwidth]{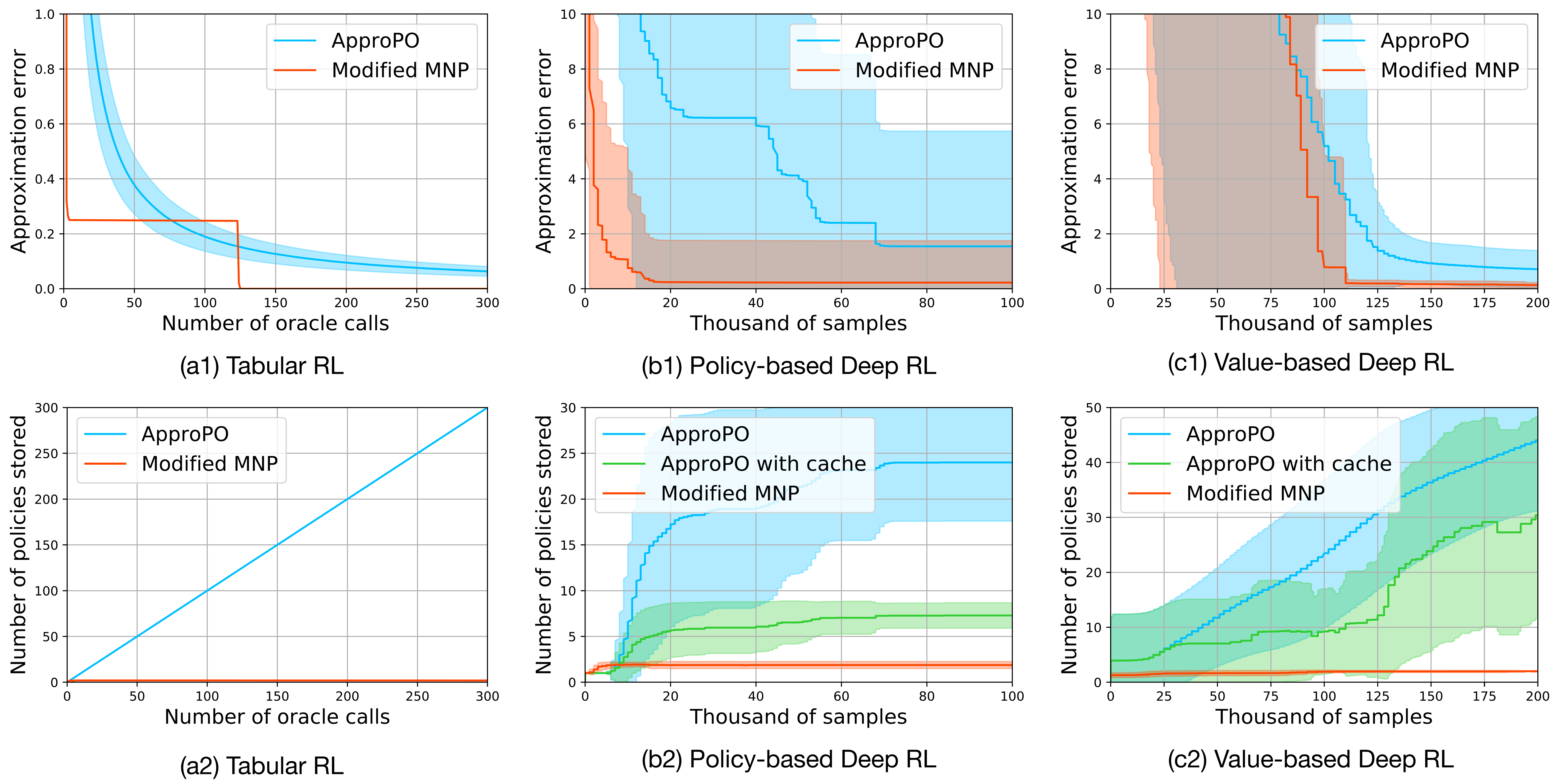}
\end{center}
\caption{The approximation error and policy efficiency of solving the navigation task using different RL methods are compared, where tabular RL (a1, a2), policy-based deep RL (b1, b2) and value-based deep RL (c1, c2) have been considered. In all three cases, our modified MNP outperforms ApproPO, and meanwhile achieves a significant memory improvement.} \label{fig:line1}
\end{figure*}

% % \vspace{-0.5em}
% \begin{figure*}[t] 
% \begin{center}
% \includegraphics[width=\textwidth]{mmnp2.pdf}
% \end{center}
% % \setlength{\textfloatsep}{1\baselineskip plus 0.2\baselineskip minus 0.5\baselineskip}
% \caption{Using a valued-based deep RL algorithm to construct the RL oracle, our method outperforms ApproPO (\textit{Left}). Meanwhile, our method stores less policies/neural networks and hence cuts the memory costs by an order of magnitudes (\textit{Middle}). A closer inspection shows that our method stores no more than $2$ policies throughout the learning process (with a guarantee of no more than $3$).}
% % \captionsetup{belowskip=0pt}
% % \vspace{-1.5em}
% \end{figure*}

\begin{theorem} [Memory Complexity Bound] \label{optimal_constant_general}
When the candidate policy set is the set of all deterministic policies, to solve CRL problems (\ref{constrained_problem}) with $m$-dimensional measurement vectors, a mixed policy needs to randomize among $m+1$ policies to ensure convergence in the worst case.
\end{theorem}
Since the minor cycles of the modified MNP method (Algorithm \ref{algo_modified_mnp}) maintain the active set to be affinely independent, the modified MNP method requires storing no more than $m+1$ individual policy, throughout the learning process. 
\begin{corollary} \label{optimal_constant}
The modified MNP method achieves the worst-case optimal policy efficiency.
\end{corollary}
Therefore we conclude that the proposed modified MNP method matches the $O(1/t)$ convergence rate of the previous game-theoretical methods.  Meanwhile, it achieves optimal policy efficiency, making it favorable for solving constrained deep RL problems.

\section{Experiments} \label{experiment_sec}

We verify the effectiveness and the efficiency of the proposed methods in a navigation task and compared them with the ApproPO \cite{miryoosefi2019reinforcement}, a game-theoretic reduction approach, using various RL methods. The ApproPO constructs an RL player similar to our RL oracle. Hence it is a natural baseline for comparison. We run experiments with the RL oracle constructed using tabular RL, policy-based deep RL, and value-based deep RL methods. In all three cases, our method outperforms ApproPO and meanwhile achieves a significant improvement in policy efficiency.

In this navigation task (Figure \ref{figure_env}), the agent is required to find a path from the starting point (S) to the goal point (G), by moving to one of the four neighborhood cells at each step. We set part of the region as risky states (grey hatch) and should be avoided. By design, the risky region contains the shortest path from S to G, so that the agent has to trade-off between a shorter path and a safer path. The agent receives a 2-dimensional measurement vector that signals the number of steps and steps inside the risky region, i.e. $(1, 0)$ for every step outside the risky region, and $(1, 1)$ for every step inside the risky region. The agent is required to find a navigation policy whose measurement vector lies inside $\Omega := \{(i,j) | 0\le i\le 11, 0\le j\le 0.5\}$. That is, the agent is required to find a policy navigating from S to G, on average containing no more than $11$ steps, and enter the risky region no more than $0.5$ steps for each episode. The episodes terminate when the goal point is reached or after $500$ steps. To simplify the presentation, we take discount  $\gamma =1$ for this finite horizon task. See Appendix C for more experimental details and hyperparameters.

A quick inspection of this task shows that none of the deterministic policies is feasible. For example, the arrows in Figure \ref{figure_env} show a deterministic policy achieving $(12,0)$ by bypassing all risk regions and one achieves $(10,1)$ by entering the risky region once. A mixed policy that randomizes these two policies with the same probability can be feasible (illustrated by the pink arrows). 

\subsection{Tabular RL Case}

% We then compare the modifed MNP method with ApproPO on this task. ApproPO is based on the game-theoretic approach and uses online graident descent (OGD) \cite{zinkevich2003online} as the online learner. Following their paper, the learning rate of OGD is set to $\eta_t := 1/\sqrt{t}$, and we initialize $\vlambda_1 = \vzero$. For the modified MNP method, we initialize $\vx_0 = \vzero$. To construct a $\mathtt{RL\_oracle}(\vlambda)$ for this tabular problem, we use dynamic programming to find an optimal deterministic policy for any given $\vlambda$. 

We first construct an RL oracle using the tabular Q-learning method. The approximation error and policy efficiency are compared in Figure 2 (a1 and a2). For the modified MNP, the method got stuck for about 100 steps. This is caused by the added projection step. As we have mentioned, a changed projected point $\vomega_t$ may yield a lower distance for the same active set $S$, and the active set remains unchanged for many steps. However, once an improving policy is found outside this active set, the modified MNP method quickly achieves the optimal value. On the other side, since the game-theoretic method gives $1/t$ weights for policies in all steps, the ApproPO slows down when getting closer to a feasible policy. For the policy efficiency (Figure 2 a2), the number of policies stored for ApproPO is simply linear to the number of oracle calls.

\subsection{Policy-based Deep RL Case}
In an online setting, we solve the navigation task using an RL oracle constructed by a deep Advantage Actor-Critic (A2C) algorithm \cite{sutton2018reinforcement,mnih2016asynchronous}. In this experiment, all methods use the same A2C agent.  ApproPO introduces extra hyper-parameters, which are set according to their original paper (see Appendix C for details), meanwhile, the proposed modified MNP introduces no extra hyper-parameters.

In Figure 2 (b1 and b2), we plot the mean and standard deviation of the approximation error and policy efficiency (number of policy stored) of running modified MNP and ApproPO methods over 50 runs. The original paper of ApproPO suggests the usages of a cache, which heuristically cuts memory costs, and does not affect its convergence. We include them in b2 and c2.

The experimental results show that our modified MNP outperforms ApproPO and meanwhile cut the memory usage by an order of magnitude. Even the memory requirement of ApproPO with cache stores significantly more policies than our proposed method. Our method stores about 2 policies throughout the process, with a guarantee of no more than 3.

\subsection{Value-based Deep RL Case}

We then consider the value-based deep RL methods, which are especially popular in offline RL settings \cite{levine2020offline,fujimoto2019off,fujimoto2019benchmarking}. We illustrate how our proposed method enables leveraging the value-based deep RL method to solve CRL tasks with the following experiments.

We first randomly collect $200$ thousand samples from the training process of the previous A2C agent, and construct a replay buffer \cite{mnih2013playing} with these samples. Then we use a Double DQN (DDQN) with dueling network \cite{wang2016dueling} to learn from samples in this replay buffer only, without any further interacting with the environment.

Learning from offline data without any further exploration is harder than in the online setting. Hence we double the training samples. Similar to our result with the policy-based RL method, when using the value-based RL method, it is clear that our proposed method also achieves superior performance. Meanwhile, throughout the learning process, our method stores much fewer policies than the ApproPO.

\section{Conclusions}

In this paper, we propose a policy efficient reduction approach to solve the CRL problem. Using a novel vector space reduction, we derive a meta-algorithm, which can be admitted by any CG-type algorithm and any RL algorithm as subroutines. To improve policy efficiency, we proposed a new variant of the CG method, the modified MNP method. The proposed method matches the $O(1/t)$ convergence rate of the existing game-theoretic methods and reduces the memory complexity from $O(1/t)$ to at most $m+1$, which is worst-case optimal. Experiments demonstrate the superior performance of our method. When working with deep RL methods, our method even cut the memory costs by an order of magnitude, making it practical to utilize deep value-based methods to solve CRL problems.

\bibliography{example_paper}
\bibliographystyle{icml2021}

\appendix

\onecolumn

\icmltitle{Appendix}

The Appendix is organized as follows. In Section A, more background on the vanilla conditional gradient (CG) method and its variant, Wolfe's method for minimum norm point (MNP), are given. Section B gives proofs of our main results. Section \ref{params_appendix} provides additional details of the aforementioned experiments.

\section{More on Conditional Gradient Type Methods}

\subsection{Vanilla Conditional Gradient}  \label{fw_appendix}

\begin{algorithm}[tbh]
\caption{Vanilla Conditional Gradient \citep{frank1956algorithm,jaggi2013revisiting}} \label{fw_algo}
\textbf{Input:}  $\mathtt{Oracle}(\cdot)$, learning rate $\{\eta_t\in[0,1]\}_{t=1,\dots,T}$\\
\textbf{Initialize}: $\vx_0\in\R^m$ 

\begin{algorithmic}[1] %[1] enables line numbers
\FOR{t = 1, \dots, T}
% \STATE $\pi_t, \vc(\pi_t) \gets \mathtt{RL\_oracle}(\vx_{t-1} - \proj_\Omega(\vx_{t-1}))$
% \STATE $\vx_t \gets (1-\eta_{t})\vx_{t-1} + \eta_t\vc(\pi_t)$, \\ 
% $\mu_t \gets (1-\eta_{t})\mu_{t-1} + \eta_t\pi_t$ \label{fw_algo_line2}

\STATE $\vs_t \gets \mathtt{Oracle}(\nabla f(\vx_{t-1}))$ \label{fw_algo_line1} 
\STATE $\vx_t \gets (1-\eta_{t})\vx_{t-1} + \eta_t\vs_t$
\ENDFOR

\STATE \textbf{return} $\vx_T$
\end{algorithmic}
\end{algorithm}

For a convex function $f: \sC \mapsto \R$, the vanilla CG method (also known as the Frank-Wolfe method) solves the constrained optimization problem $\min_{\vx\in\sC} f(x)$ over a compact and convex set $\sC$ using a linear optimization oracle $\mathtt{Oracle}(\vv) := \argmin_{\vs\in\sC}\vs^T\vv$. The process is illustrated in Algorithm 3. For $\eta_t := \frac{2}{t+1}$, the vanilla CG is known to have a sublinear $O(1/t)$ convergence rate \cite{jaggi2013revisiting}. Various methods are proposed to improve the convergence rate. For example, when $\sC$ is a polytope, and the objective function is strongly convex, multiple variants, such as away-step CG \cite{wolfe1970convergence,jaggi2013revisiting}, pairwise CG \cite{mitchell1974finding}, and Wolfe's method \cite{wolfe1976finding} are shown to enjoy linear convergence rate \cite{lacoste2015global}. Linear convergence under other conditions is also studied \cite{beck2017linearly,garber2013linearly,garber2013playing}.

\subsection{Wolfe's Method for Minimum Norm Point} \label{wolfe_appendix}

\begin{algorithm} [h]
\caption{Wolfe's Method for Minimum Norm Point \cite{wolfe1976finding}}  \label{wolfe_method_mnp}
\textbf{Input:}  $\mathtt{Oracle}(\cdot)$ \\
\textbf{Initialize}: current point $\vx_0\in\sP$, active set $\sS = \{\}$, weight $\mu(\cdot)=0$ for points in $\sS$.

\begin{algorithmic}[1] %[1] enables line numbers
% \WHILE[Major cycle]{true}
\FOR[Major cycle]{t = 1, \dots, T}
\STATE $\vs_t \gets \mathtt{Oracle}(\vx_{t-1})$ \COMMENT{Potential improving point}
% \STATE \textbf{if} $||\vx||^2 \le \vx^T\vs + \epsilon$ \textbf{then} break
\STATE $\sS \gets \sS \cup \{\vs_t\}$
\WHILE[Minor cycle]{True} 
  \STATE $\vy, \alpha \gets \mathtt{AffineMinimizer}(\sS)$
  \IF[$\sS$ is a corral]{$\forall \vs \in \sS, \alpha(\vs) > 0$}
  \STATE \textbf{break}
  \ELSE
    \STATE $\theta \gets \min_{\vs: \alpha(\vs)\le 0} \frac{\mu(\vs)}{\mu(\vs) - \alpha(\vs)}$
    \STATE $\vx_{t-1} \gets \theta\vy + (1-\theta) \vx_{t-1}$ \\$ \mu(\vs) \gets \theta\alpha(\vs) + (1-\theta) \mu(\vs)$ for all $\vs\in\sS$
    \STATE $\sS \gets \{\vs\in\sS\ |\ \mu(\vs)>0\}$
  \ENDIF
\ENDWHILE
\STATE $\vx_t \gets \vy, \vmu \gets \valpha$
\ENDFOR
\STATE \textbf{return} $\vx_T$

\end{algorithmic}
\end{algorithm}

Wolfe's method for minmum norm point (MNP) problem is an iterative algorithm to find the point with minimum Euclidean norm in a polytope, where the polytope is defined as the convex hull of a set of finitely many points $\sP := \conv(\{\vs_1, \dots, \vs_n\})$. The Wolfe's method consists of a finite number of major cycles, each of which consists of a finite number of minor cycles. The original MNP method iterates until a termination criteria is satisfied. At the start of each major cycle, let $H(\vx) := \{\vy^T\vx = \vx^T\vx | \vx \in\R^m\}$ be the hyperplane defined by $\vx$. If $H(\vx)$ separates the polytope from the origin, then the process is terminated. Otherwise, it invokes an oracle to find any point on the near side of the hyperplane. The point is then added into the active set $\sS$, and starts a minor cycle. 

In a minor cycle, let $\vy$ be the point of smallest norm in of the affine hull $\aff(\sS)$. If $\vy$ is in the relative interior of the convex hull $\conv(\sS)$, then $\vx$ is updated to $\vy$ and the minor cycle is terminated. Otherwise, $\vy$ is updated to the nearest point to $\vy$ on the line segment $\conv(\sS) \cap [\vx_{t-1}, \vy]$. Thus $\vy$ is updated to a boundary point of $\conv(\sS)$, and any point that is not on the face of $\conv(\sS)$ in which $\vy$ lies is deleted. The minor cycles are executed repeatedly until $\sS$ becomes a \textit{corral}, that is, a set whose affine minimizer lies inside its convex hull. Since a set of one point is always a corral, the minor cycles is terminated after a finite number of runs. 

\section{Proofs of the Main Results}

Recall that $\vx_t := \vc(\mu_t) $ (measurement of the mixed policy) throughout the process. In the following proofs, we define $\vs_t := \vc(\pi_t)$ (measurement of the latest found policy) to simplify notation. When discussing one major cycle step with $t$ fixed, let $\vy_i$ denotes the affine minimizer found in the $i$-th minor cycle (line 6 of Algorithm 2).

\subsection{Proof of Proposition 4.1}
\begingroup
\def\thetheorem{4.1}
\begin{proposition}
For any convex set $\Omega$, the function $f(\vx) := \frac{1}{2} \dist^2(\vx, \Omega)$ is strongly convex if and only if $\Omega$ is a singleton. 
\end{proposition}
\addtocounter{theorem}{-1}
\endgroup

Recall that a function $f$ over $S$ is defined to be strongly convex  \cite{boyd2004convex}. If $\exists m>0$, such that $\forall \vu,\vv \in S$
\begin{align*}
    f(\vu)\ge f(\vv) + \nabla f(\vv)^T(\vu-\vv) + \frac{m}{2}||\vu-\vv||^2.
\end{align*}

\begin{proof}
"If" part: when $\Omega$ is a singleton, the target function $f(\cdot)$ is twice continuously differentiable, with $\nabla^2f(\vx) = 1$, and hence is strongly convex with $m=1$. The “only if” part can be proved by contrapositive. For a non-singleton convex set, taking two distinct points from the set, any convex combination of them achieves 0 for $f$, i.e., $f$ is not strictly convex, and hence not strongly convex. 
\end{proof}

\subsection{Proof of Theorem 4.2} 

The idea is to consider the distance between $\vx_t$ and $\vomega_t$. When the major cycle has no minor cycle, the non-terminal condition and the affine minimizer property implies $\dist^2(\vx_{t+1}, \vomega_t) < \dist^2(\vx_t, \vomega_t)$. Otherwise we show that the first minor cycle strictly reduces the $\dist^2(\vx_t, \vomega_t)$ by moving along the segment joining $\vx$ and $\vy$, and the subsequent minor cycle cannot increase it. Since $\vomega_t\in\Omega$, we conclude $\err(\vx_{t+1}) \le \dist^2(\vx_{t+1},\omega_t) < \dist^2(\vx_t,\omega_t) = \err(\vx_{t})$, and the approximation error strictly decreases.

\begingroup
\def\thetheorem{4.2}
\begin{theorem} [Approximation Error Strictly Decreases] 
For each $t>1$ step, the $\vx_t, \mu_t$ found by Algorithm 2 satisfies $\err(\mu_{t}) < \err(\mu_{t-1})$. That is, the measurement vectors of $\mu_{t}$ gets strictly closer to the convex set $\Omega$.
\end{theorem}
\addtocounter{theorem}{-1}
\endgroup

\begin{proof}
If the current step is a major cycle with no minor cycle, then $\vx_{t+1}$ is the affine minimizer of $\aff(\sS \cup \{\vs_t\})$ with respect to $\vomega_t$. Then the affine minimizer property implies $(\vs_t - \vx_{t+1}) (\vx_{t+1} -\omega_t) = 0$. Since iteration does not terminate at step $t$, we have $(\vx_t-\vomega_t)^T(\vx_t-\vs_t)>0$ (Wolfe's Criterion \cite{wolfe1976finding}), and therefore $\vx_{t+1}$ not equal to $\vx_t$. Then $\vx_{t+1}$ is the unique affine minimizer implies $f(\vx_{t+1}) = \min_{\omega\in\Omega} ||\vx_{t+1}-\omega||^2 \le ||\vx_{t+1}-\omega_t||^2 < ||\vx_{t}-\omega_t||^2 = f(\vx_t)$.

Otherwise the current step contains one or more minor cycles. In this case, we show that the first minor cycle strictly reduces the approximation error, and the (possibly) following minor cycles cannot increase it. For the first minor cycle, the affine minimizer $\vy_0$ of $\aff(\sS \cup \{\vs_t\})$ with respect to $\vomega_t$ is outside $\conv(\sS \cup \{\vs_t\})$. Let $\vz = \theta \vy_0 + (1-\theta)\vx_t$ be the intersection of $\conv(\sS \cup \{\vs_t\})$ and segment joining $\vx$ and $\vy$. Let $\sV_0 := \sS_t$ and $\sV_i$ denote the active set after the $i$-th minor cycle. Then since $\vy_1$ is the affine minimizer of $\sV_1$ with respect to $\omega_t$, we have
\begin{align}
||\vz - \omega_t|| =  ||\theta \vy_0 + (1-\theta)\vx_t-\omega_t|| \le \theta ||\vy_0 -\omega_t|| + (1-\theta)||\vx_t-\omega_t|| < ||\vx_t-\omega_t||, \label{thm_strictly_decrease_eq1}
\end{align}
where the second step uses the triangle inequality and the last step follows since the segment $\vx_t \vy_0$ intersects the interior of $\conv(\sS \cup \{\vs_t\})$, and the distance to $\omega_t$ strictly decreases along this segment. Therefore the point $\vy_1$ found by first minor cycle satisfies
% \begin{align}
%     \theta ||\vy_0 -\omega_t|| + (1-\theta)||\vx_t-\omega_t|| .
% \end{align}
\begin{align}
    f(\vy_1) = \min_{\omega\in\Omega}||\vz - \omega||^2\le ||\vz - \omega_t||^2 < ||\vx_t-\omega_t|| = f(\vx_t).
\end{align}
Minus both side by the optimal value of the problem $f(\vx^*)$, it it clear that the first minor cycle strictly decreases the approximation error. By a similar argument, in subsequent minor cycles the approximation error cannot be increased. However, after the first minor cycle, the iterating point may already at the intersection point and the strict inequality in last step of Eq. (\ref{thm_strictly_decrease_eq1}) need to be replaced by non-strict inequality.

Therefore any major cycle either finds an improving point and continue, or enters minor cycles where the first minor cycle finds an improving point, and the subsequent minor cycles does not increase the distance. Adding both side of $f(\vx_{t+1}) < f(\vx_{t})$ by $f(\vx^*)$ and we have the approximation error $\err(\mu_t)$ strictly decreases.
\end{proof}

% To prove the Lemma \ref{lemma_non_drop_steps}, we use a proposition from \citet{chakrabarty2014provable}.

\subsection{Proof of Theorem 4.3} \label{convergence_proof}

In our analysis, we consider the approximation error as defined in (4)
\begin{align*}
\mathtt{err}(\mu) := \frac{1}{2}\dist^2(\vc(\mu), \Omega) - \frac{1}{2}\dist^2(\vc(\mu^*), \Omega).
\end{align*}
We first prove the following Lemma \ref{lemma_non_drop_steps} and Lemma \ref{lemma_drop_steps}. Then we present the proof of Theorem 4.3 using the lemmas. 

% The main techniques are based on \citet{chakrabarty2014provable}. However, our Lemma \ref{lemma_drop_steps} give a tighter bound on the frequency of bad steps, and hence we improve the convergence rate from their $O(m/t)$ to a tighter $O(1/t)$.

\begin{lemma} \label{lemma_non_drop_steps}
For a non-drop step, we have $\err(\mu_t) - \err(\mu_{t+1}) \ge \err^2(\mu_t)/8Q^2$.
\end{lemma}

\begin{proof}

The non-drop step contains either no minor cycle or one minor cycle. We first consider the no minor cycle case.

If a major cycle contains no minor cycle, then $\vx_{t+1}$ is the affine minimizer of the $\sS \cup \{\vs_t\}$.

% (Theorem \ref{}) For a iteration $t$ of major cycle with no minor cycles, we show that $h(\vx_{t+1}) - h(\vx_t) \ge \frac{g(\vx_t)^2}{4Q^2}$, where $g(\vx) := \max_{\vs\in\sC} (\vx-\vs)^T(\vx-\proj_\Omega(\vx))$ and $Q$ is the upper bound on $||\vx - \proj_\Omega(\vx)||$.
\begin{align}
    \err(\mu_t) - \err(\mu_{t+1}) &= \dist^2(\vx_t, \Omega) - \dist^2(\vx_{t+1}, \Omega) \\
        & = 1/2 (||\vx_t-\vomega_t||^2 - \min_{\vomega\in\Omega}||\vx_{t+1} - \vomega||^2) \label{no_minor_eq_2}\\
        & \ge 1/2 (||\vx_t-\vomega_t||^2 - ||\vx_{t+1}-\vomega_t||^2) \label{no_minor_eq_3}\\
        & = 1/2 (||\vx_t-\vomega_t||^2 + ||\vx_{t+1}-\vomega_t||^2 - 2 ||\vx_{t+1}-\vomega_t||^2) \label{no_minor_eq_4}\\
        & = 1/2 (||\vx_t-\vomega_t||^2 + ||\vx_{t+1}-\vomega_t||^2 - 2 (\vx_t-\vomega_t)^T(\vx_{t+1}-\vomega_t)) \label{no_minor_eq_5} \\
        & = 1/2 (||\vx_{t}-\vx_{t+1}||^2), \label{no_minor_eq_6}
\end{align}
where the equation (\ref{no_minor_eq_5}) follows from the affine minimizer property Eq. (9). For $||\vx_{t}-\vx_{t+1}||$ in the last equation, and $\forall \vq \in \aff(\sS \cup \{\vs_t\})$, we have
\begin{align}
    ||\vx_{t}-\vx_{t+1}|| &\ge ||\vx_{t}-\vx_{t+1}|| \frac{||\vx_{t}|| + ||\vq||}{2Q} &  (\text{ Definition of } Q)\\
        &\ge ||\vx_{t}-\vx_{t+1}|| \frac{||\vx_{t} - \vq||}{2Q} & (\text{ Triangle inequality})\\
        &\ge \frac{1}{2Q} (\vx_{t}-\vx_{t+1})(\vx_{t} - \vq) & (\text{ Cauchy-Schwarz inequality})\\
        &= \frac{1}{2Q} (\vx_{t}-\vomega_t)(\vx_{t} - \vq) & (\text{ Affine minimizer property}). \label{no_minor_eq_7}
\end{align}
Then it suffices to show that  $(\vx_{t}-\vomega_t)(\vx_{t} - \vq) \ge \err(\mu_t)$. 

Since $\Omega$ is a convex set, the squared Euclidean distance function $\dist^2(\vx, \Omega)$ is convex for $\vx$, which implies 
\begin{align}
    \dist^2(\vx_t, \Omega) + (\vq-\vx_t)\nabla \dist^2(\vx_t, \Omega) \le \dist^2(\vq, \Omega). \label{no_minor_eq_8}
\end{align}
Putting in $\nabla \dist^2(\vx_t, \Omega) = (\vx_t - \proj_\Omega(\vx_t)) = (\vx_t - \vomega_t)$, we get $(\vx_{t}-\vomega_t)(\vx_{t} - \vq) \ge \err(\mu_t)$, which together with Eq. \ref{no_minor_eq_6} and Eq. \ref{no_minor_eq_7} concludes that for non-drop step with no minor cycles, we have $\err(\mu_t) - \err(\mu_{t+1}) \ge \err^2(\mu_t)/8Q^2$. 

For non-drop step with one minor cycle, we use the Theorem 6 of \citep{chakrabarty2014provable}. By a linear translation of adding all points with $-\omega_t$, it gives
\begin{align}
    ||\vx_t - \vomega_t||^2 - ||\vx_{t+1} -\vomega_t||^2 \ge ((\vx_{t}-\vomega_t)(\vx_{t} - \vq) )^2/8Q^2.
\end{align}
Then applying the same argument as Eq. \ref{no_minor_eq_8}, and we finished our proof.

\end{proof}

\begin{lemma} \label{lemma_drop_steps}
After $t$ major cycle steps of modified MNP method, the number of drop steps is less than $t/2$.
\end{lemma}

%  proof of convergence
\begin{proof}

Since Lemma \ref{lemma_drop_steps} shows that drop steps are no more than half of total major cycle steps, and Theorem 4.2 guarantees these drop steps reducing the approximation error, we can safely skip these step, and re-index the step numbers to include non-drop steps only using $k$.

For these non-drop steps, we claim that $\err(\mu_k) \le 8Q^2/(k+1)$. Using Lemma \ref{lemma_non_drop_steps}, we prove the convergence rate using induction. 
We first bound the error of any $\err(\mu_k)$. For any $k\ge 1$
\begin{align}
\err(\mu_k) &= \dist^2(\vc(\mu_k), \Omega) - \dist^2(\vc(\mu^*), \Omega) \\
 &= 1/2 ||\vc(\mu_k) - \proj_\Omega(\vc(\mu_k))||^2 - 1/2 ||\vc(\mu^*) - \proj_\Omega(\vc(\mu^*))||^2 \label{induction_line1}\\
 &\le 1/2 (||\vc(\mu_k)||^2 + ||\proj_\Omega(\vc(\mu_k))||^2 - ||\vc(\mu^*)||^2 -  ||\proj_\Omega(\vc(\mu^*))||^2) \label{induction_line2}\\
 &\le ||\vc(\mu_k)||^2 - ||\vc(\mu^*)||^2 \label{induction_line3}\\
 &\le ||\vc(\mu_k)||^2 \\
 &\le Q^2, \label{induction_line4}
\end{align}
where Eq. \ref{induction_line1} uses the definition of our squared Euclidean distance function. Eq. \ref{induction_line2} follows from triangle inequality, and Eq. \ref{induction_line3} is by the contractive property of the Euclidean distance.

When $k=1$, the Eq. \ref{induction_line4} established the based case. Now for $k\ge 1$, assume that $\err(\mu_k) \le 8Q^2/(k+1)$ for $k\ge1$, then Lemma \ref{lemma_non_drop_steps} gives $\err(\mu_{k+1}) \le \err(\mu_k) - \err^2(\mu_k)/8Q^2$. Since the quadratic function of the right hand side is monotonically increasing on $(-\infty, 4Q^2]$, using the inductive hypothesis  
\begin{align}
    \err(\mu_{k+1}) \le \err(\mu_k) - \err^2(\mu_k)/8Q^2 \le  8Q^2/(k+1) -  8Q^2/(k+1)^2 \le Q^2/(k+2)
\end{align}

Since for $t$ steps of major cycle steps, the number of non-drop steps $k > t/2$, we conclude that  $\err(\mu_t) \le 16Q^2/(t+2)$.

\end{proof}

Then we prove the Theorem 4.3.
\begingroup
\def\thetheorem{4.3}
\begin{theorem} [Convergence in Approximation Error]
For $t\ge 1$, the mixed policy $\mu_t$ found by the Algorithm \ref{algo_modified_mnp} satisfies
\begin{align}
    \err(\mu_t) \le 16Q^2/(t+2),
\end{align} 
where $Q := \max_{\pi\in\Pi} ||J(\pi)||$ is the maximum norm of a measurement vector. 
\end{theorem}
\addtocounter{theorem}{-1}
\endgroup

\begin{proof}
Recall that at the termination of a minor cycle, the size of the active set $|\sS_c| \in [1,m]$. Since in each major cycle steps, the size of active set $\sS_t$ increases by one, and each drop step reduces the size of $\sS_t$ by at least one, the number of drop steps is always less than half of total number of the major cycle steps.
\end{proof}

% Having bound the fraction of drop steps, then it suffices to show that the non-drop step significantly reduces the approximation error (see Lemma \ref{lemma_non_drop_steps} in Appendix \ref{convergence_proof}), and Theorem \ref{convergence} follows

% The main techniques to prove this proposition are based on the analysis of \citet{chakrabarty2014provable} to the Wolfe's method. We include a proof in the Appendix \ref{} for completeness.

\subsection{Proof of Theorem 4.4} \label{optimal_constant_general_proof}

\begingroup
\def\thetheorem{4.4}
\begin{theorem} [Memory Complexity Bound]
For CRL problems (\ref{constrained_problem}) with $m$-dimensional measurement vectors, when deterministic policies are used, a mixed policy needs to randomize among $m+1$ policies to ensure convergence in the worst case.
\end{theorem}
\addtocounter{theorem}{-1}
\endgroup

\begin{proof}
We give a constructive proof. Consider a $m$-dimensional vector-valued MDP with a single state, $m+1$ actions, and $\vc(a_i) := \ve_i$ is the unit vector of $i$-th dimension for $i\in[1,m]$, and $\vc(a_{m+1}) := \vzero$, and the episode terminates after $1$ steps. The CRL problem is to find a policy whose measurement vector lies in the convex set of a single point $\{\vone/2m\}$. By linear programming, it is clear that the only feasible mixed deterministic policy is to select $a_{m+1}$ with $1/2$ probability, and the rest $m$ actions with $1/2m$ probability; i.e. the unique feasible policy to this problem has an active set containing $m+1$ deterministic policies. Therefore any method randomize among less than $m+1$ individual policies does not ensure convergence when used with RL algorithms searching for deterministic policies.
\end{proof}

\section{Additional Experimental Details} \label{params_appendix}
\subsection{Details of the Policy-based RL Experiments}

In this experiment, all methods use the same A2C agent, where the hyper-parameters are chosen following the ApproPO paper. The input is the one-hot encoded current position index, and both actor and critic use fully connected multi-layer perceptrons with the ReLU activation function. The actor and critic share the internal representation and have their own final layers. Both actor and critic networks use Adam optimizer with learning rate set to 1e-2 (see Table 2 for more details).

ApproPO introduces extra hyper-parameters. Following the original paper, the constant $\kappa$ for projection convex set to convex cone is set to be 20. The $\vtheta$ is initialized to 0. The proposed modified MNP introduces no extra hyper-parameters, and has nothing to report.

\begin{table}[tbh]
\caption{Network structures and parameters of the A2C algorithm used in Sec. 5.} \label{a2c_param}
\begin{center}
\begin{tabular}{|c|c|c|} \hline
 &  Actor & Critic \\
\hline
Input layer & \multicolumn{2}{c|}{One-hot encoded state index (dim=$54$)} \\
\hline
Hidden layer & \multicolumn{2}{c|}{Linear(in=$54$, out=$128$, act="relu")} \\ 
\hline
Output layer & Linear(in=$128$, out=$4$, act="relu") & Linear(in=$128$, out=$1$, act="relu") \\ 
\hline
Output name & Action score & State value \\ 
\hline
\end{tabular}
\end{center}
\end{table}

\subsection{Details of the Value-based RL Experiments}

The network structure of DDQN with Dueling network is specified in Table 3. For each iteration, we randomly samples a batch of 256 transition samples, and we synchronous the target network every 10 iterations. The same agent is used for both methods, with optimizer set to Adam with a learning rate of 1e-3.

\begin{table}[h]
\caption{Network structures and parameters of the DDQN with Dueling Network algorithm used in Sec. 5.}
\begin{center}
\begin{tabular}{|c|c|c|} \hline
 &  \multicolumn{2}{c|}{DDQN with Dueling Network} \\
\hline
Input layer & \multicolumn{2}{c|}{One-hot encoded state index (dim=$54$)} \\
\hline
Hidden layer & Linear(in=$54$, out=$128$, act="relu") & Linear(in=$54$, out=$128$, act="relu") \\ 
\hline
Output layer & Linear(in=$128$, out=$128$, act="relu") &  Linear(in=$128$, out=$128$, act="relu") \\
\hline
Output layer & Linear(in=$128$, out=$4$, act="relu") & Linear(in=$128$, out=$1$, act="relu") \\
\hline
Output name & Advantage score: $\mathtt{adv}$& State score: $\mathtt{v}$\\ 
\hline
Q value & \multicolumn{2}{c|}{$\mathtt{q} \gets \mathtt{v} + \mathtt{adv} - \mathtt{average}(\mathtt{adv}$)}\\ 
\hline
\end{tabular}
\end{center}
\end{table}

% \section{Additional Experiment Details} \label{params_appendix}

% All the methods use the same A2C agent. The input is the one-hot encoded current position index. The A2C is the standard fully connected multi-layer perceptron with ReLU activation function. The actor and critic share the internal representation and have their only final layer. Both actor and critic networks use Adam optimizer with learning rate set to $1e^{-2}$. The network is as follows

% \begin{table}[!hp]
% \begin{center}
% \begin{tabular}{|c|c|c|} \hline
%  &  Actor & Critic \\
% \hline
% Input layer & \multicolumn{2}{c|}{One-hot encoded state index (dim=$54$)} \\
% \hline
% Hidden layer & \multicolumn{2}{c|}{Linear(in=$54$, out=$128$, act="relu")} \\ 
% \hline
% Output layer & Linear(in=$128$, out=$4$, act="relu") & Linear(in=$128$, out=$1$, act="relu") \\ 
% \hline
% Output name & Action score & State value \\ 
% \hline
% \end{tabular}
% \end{center}
% \end{table}
% For ApproPO, the constant $\kappa$ for projection convex set to convex cone is set to be $20$. The $\vtheta$ is initialized to $0$. Following the original paper. 

% For RCPO, the learning rate of its $\vlambda$ is set to $2.5e^{-5}$, and its $\vlambda$  is initialized to $0$ and updated by online gradient descent with learning rate set to $1$, as used by the original paper. 

% The proposed modified MNP introduces no extra hyper-parameters, and has nothing to report.

\clearpage

\end{document}